\documentclass[sigconf]{acmart}

\usepackage{graphicx}
\usepackage{float}
\usepackage{verbatim}
\usepackage{tikz}
\usepackage{pgfplots}
\usepackage{xcolor} 
\usetikzlibrary{shapes.geometric, arrows}
\usepackage{array,ragged2e}
\newcolumntype{P}[1]{>{\RaggedRight\arraybackslash}p{#1}}
\usepackage[compact]{titlesec}
\usepackage{tabularx}
\usepackage{balance} 

\AtBeginDocument{%
  }

\copyrightyear{2025}
\acmYear{2025}
\setcopyright{cc}
\setcctype{by}
\acmConference[KDD '25]{Proceedings of the 31st ACM SIGKDD Conference on Knowledge Discovery and Data Mining V.2}{August 3--7, 2025}{Toronto, ON, Canada}
\acmBooktitle{Proceedings of the 31st ACM SIGKDD Conference on Knowledge Discovery and Data Mining V.2 (KDD '25), August 3--7, 2025, Toronto, ON, Canada}
\acmDOI{10.1145/3711896.3736570}
\acmISBN{979-8-4007-1454-2/2025/08}

\begin{document}

\title{Evaluation and Benchmarking of LLM Agents: A Survey}

\author{Mahmoud Mohammadi}
\email{mahmoud.mohammadi@sap.com}
 \orcid{0000-0001-6829-1420}
\affiliation{%
  \institution{SAP Labs}
  \city{Bellevue}
  \state{WA}
  \country{USA}
}

\author{Yipeng Li}
\email{yipeng.li@sap.com}
\orcid{0009-0002-4076-864X}
\affiliation{%
  \institution{SAP Labs}
  \city{Bellevue}
  \state{WA}
  \country{USA}
}

\author{Jane Lo}
\email{jane.lo@sap.com}
\orcid{0009-0007-6117-5022}
\affiliation{%
  \institution{SAP Labs}
  \city{Palo Alto}
  \state{CA}
  \country{USA}
}

\author{Wendy Yip}
\email{wendy.yip@sap.com}
\orcid{0009-0008-1734-2935}
\affiliation{%
  \institution{SAP Labs}
  \city{Palo Alto}
  \state{CA}
  \country{USA}
}



\begin{abstract}
The rise of LLM-based agents has opened new frontiers in AI applications, yet evaluating these agents remains a complex and underdeveloped area. This survey provides an in-depth overview of the emerging field of LLM agent evaluation, introducing a two-dimensional taxonomy that organizes existing work along (1) evaluation objectives—what to evaluate, such as agent behavior, capabilities, reliability, and safety—and (2) evaluation process—how to evaluate, including interaction modes, datasets and benchmarks, metric computation methods, and tooling. In addition to taxonomy, we highlight enterprise-specific challenges, such as role-based access to data, the need for reliability guarantees, dynamic and long-horizon interactions, and compliance, which are often overlooked in current research.  We also identify the future research directions, including holistic, more realistic, and scalable evaluation. This work aims to bring clarity to the fragmented landscape of agent evaluation and provide a framework for systematic assessment, enabling researchers and practitioners to evaluate LLM agents for real-world deployment.
\end{abstract}



\maketitle

\section{Introduction}

Agents based on LLMs are autonomous or semi-autonomous systems that use LLMs to reason, plan, and act, and represent a rapidly growing frontier in artificial intelligence \cite{Yao2023, nakajima2023babyagi}. From customer service bots and coding copilots to digital assistants, LLM agents are redefining how we build intelligent systems. 

As these agents move from research prototypes to real-world applications \cite{fourney_magentic-one_2024, lu2024ai}, the question of how to rigorously evaluate them becomes both pressing and complex.
However, evaluating LLM agents is more complex than evaluating LLMs in isolation. Unlike LLMs, which are primarily assessed for text generation or question answering, LLM agents operate in dynamic, interactive environments. They reason and make plans, execute tools, leverage memory, and even collaborate with humans or other agents \cite{durante2024agent}. This complex behavior and dependence on real-world effects make standard LLM evaluation approaches insufficient. To make an analogy, LLM evaluation is like examining the performance of an engine. In contrast, agent evaluation assesses a car's performance comprehensively, as well as under various driving conditions.

LLM agent evaluation also differs from traditional software evaluation. While software testing focuses on deterministic and static behavior, LLM agents are inherently probabilistic and behave dynamically; therefore, they require new approaches to assessing their performance. The evaluation of LLM agents is at the intersection of natural language processing (NLP), human-computer interaction (HCI), and software engineering, which demands additional perspectives.

Despite increasing interest in this space, existing surveys focus narrowly on LLM evaluation or cover specific agent capabilities without a holistic perspective \cite{zhang_survey_2024}. In addition, enterprise applications bring additional requirements to agents, including secure access to data and systems, a high degree of reliability for audit and compliance purposes, and more complex interaction patterns, which are rarely addressed in the existing literature \cite{yehudai_survey_2025}. This survey aims to serve as a helpful reference for practitioners and researchers in the field of agent evaluation. Our contributions in this survey are twofold. 
\begin{itemize}
    \item We propose a taxonomy of LLM agent evaluation that organizes prior work by evaluation objectives (what to evaluate, such as behavior, capabilities, reliability, and safety) and evaluation process (how to evaluate, including interaction modes, datasets and benchmarks, metrics computation methods, evaluation tooling, and evaluation environments). 
    \item We highlight enterprise-specific challenges, including role-based access control, reliability guarantees, long-term interaction, and compliance requirements. 
\end{itemize}

The remainder of this paper is structured as follows. Section 2 describes the taxonomy used in this survey paper to analyze the agent evaluation landscape. Section 3 dives into the first dimension of the taxonomy, evaluation objectives, and focuses on the aspects of the agent to be evaluated. Section 4 describes the second dimension, the evaluation process, and focuses on the evaluation method. Section 5 discusses the challenges of assessing LLM agents in enterprise environments. Section 6 outlines open questions and future research directions to guide the next phase of work in evaluating LLM agents. 

\section{Taxonomy for LLM-based Agent Evaluation}
\begin{figure*}
    \centering
    \includegraphics[width=\textwidth]{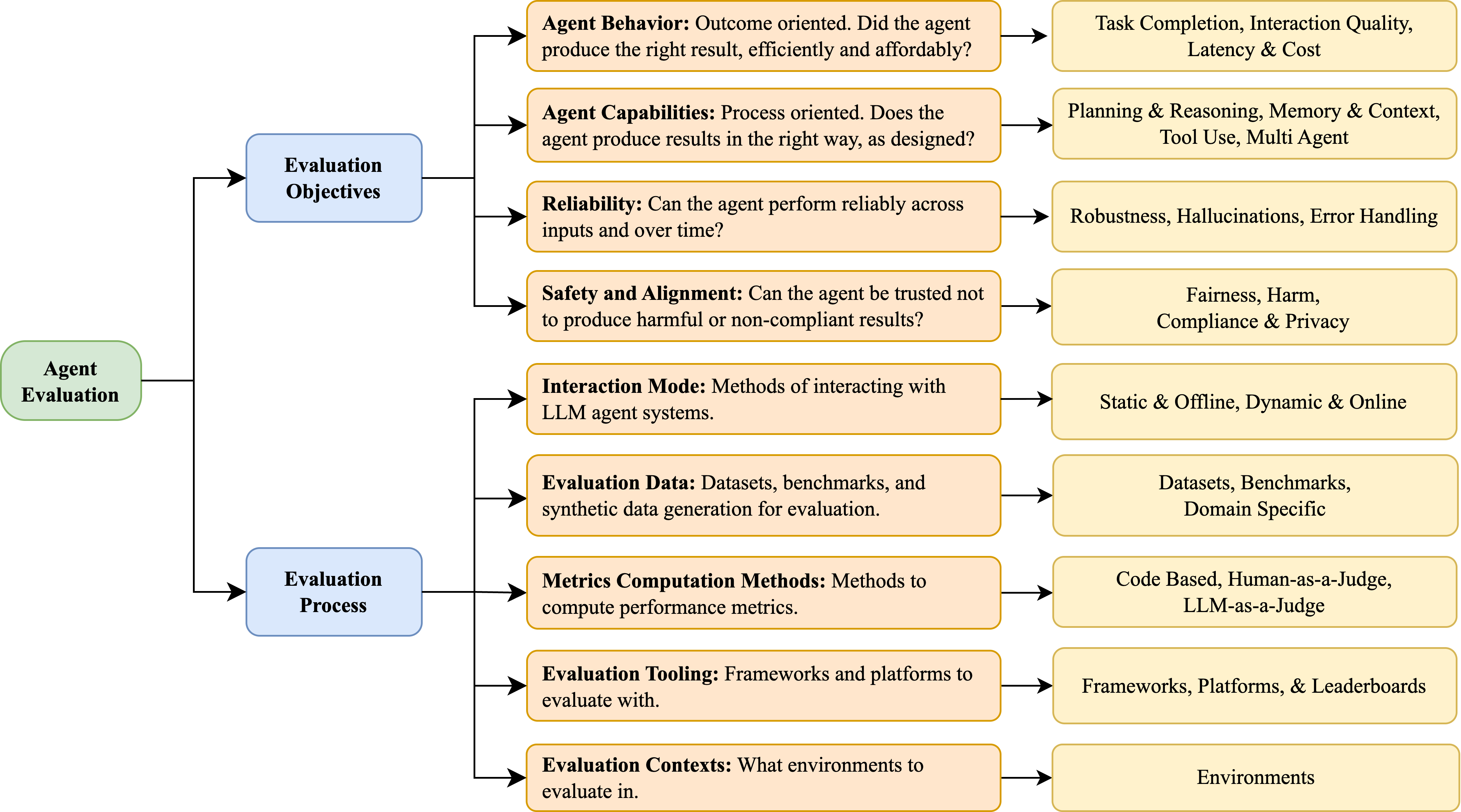}
    \caption{Taxonomy of LLM Agent Evaluation}
    \label{fig:taxonomy}
\end{figure*}

We propose a two-dimensional taxonomy to organize different aspects of the evaluation of LLM-based agents, structured along the axes of \textbf{Evaluation Objectives} (what to evaluate) and \textbf{Evaluation Process} (how to evaluate). This taxonomy is visualized as a hierarchical tree in \ref{fig:taxonomy}.

The \textit{Evaluation Objectives} dimension is concerned with the targets of evaluation. The first category, \textit{Agent Behavior}, in this dimension focuses on outcome-oriented aspects such as task completion and output quality, capturing how well an agent meets end-users' expectations. Next, \textit{Agent Capabilities} emphasize process-oriented competencies, including tool use, planning and reasoning, memory and context retention, and multi-agent collaboration. These capabilities provide insights into how agents achieve their goals and how well they meet their design specification. \textit{Reliability} assesses whether an agent behaves consistently for the same input and robustly when input varies or the system encounters errors. Finally, \textit{Safety and Alignment} evaluates the agent’s trustworthiness and security, including fairness, compliance, and the prevention of harmful or unethical behaviors.

The \textit{Evaluation Process} dimension describes how agents are assessed. \textit{Interaction Mode} distinguishes between static evaluation, where agents respond to fixed inputs, and interactive assessment, where agents engage with users. \textit{Evaluation Data} discusses both synthetic and real-world datasets, as well as benchmarks tailored to specific domains such as software engineering, healthcare, and finance \cite{fourney_magentic-one_2024, lu2024ai}. \textit{Metrics Computation Methods} encompasses quantitative measures, such as task success and factual accuracy, as well as qualitative evaluations based on human or LLM judgments. \textit{Evaluation Tooling} refers to the supporting infrastructure, such as instrumentation frameworks (e.g., LangSmith, Arize AI) and public leaderboards (e.g., Holistic Evaluation of Agents), that enable scalable and reproducible assessment. Lastly, \textit{Evaluation Contexts} define the environment in which evaluations are conducted, from controlled simulations to open-world settings such as web browsers or APIs.

This taxonomy serves both as a conceptual framework and a practical guide, enabling systematic comparison and analysis of LLM agents across a wide range of goals, methodologies, and deployment conditions. In the following sections, we examine each dimension in detail, highlighting key evaluation practices and representative studies.

As LLM agents are deployed in increasingly diverse and complex settings, factors such as single-turn versus multi-turn interactions, multilingualism, and multimodality all become more important. While the taxonomy remains applicable across these variations, tailored metrics and evaluation strategies are often required. We will discuss these specific adaptations in the relevant sections that follow.

\section{Evaluation Objectives} \label{sec:eval_objectives}

\subsection{Agent Behavior}
Agent behavior refers to the overall performance of the agent as perceived by a user, treating the agent as a black box. It represents the highest-level view in evaluation and offers the most direct insight into the user experience. This category encompasses aspects such as task completion, output quality, latency, and cost.

\subsubsection{Task Completion:} 
Task completion is a fundamental objective of agent evaluation, assessing whether an agent successfully achieves the predefined goals of a given task \cite {reimann_predicting_2023, chen_scienceagentbench_2025,xiao_flowbench_2024, zhang_large_2025}. It involves determining whether a desired state is reached or if specific criteria defined for task success are met \cite{liu_agentbench_2023, wang_survey_2024}. Although sometimes noted for providing limited fine-grained insight into failures, especially when most models achieve low success rates \cite{ma_agentboard_2024}, task completion remains a predominant and essential measure of overall agent performance \cite{ma_agentboard_2024}.

Task completion is commonly quantified using metrics such as \textit{Success Rate} (SR) \cite{liu_agentbench_2023, ma_agentboard_2024, chen_scienceagentbench_2025}, which can also be referred to as \textit{Task Success Rate} \cite{zhang_large_2025} or \textit{Overall Success Rate} \cite{liu_agentbench_2023}. Other related metrics include \textit{Task Goal Completion} (TGC) \cite{trivedi_appworld_2024} and \textit{Pass Rate} \cite{qin_toolllm_2023}. Some evaluations employ binary indicators, such as a reward function that returns 0 or 1 for goal achievement \cite{koh_visualwebarena_2024}. 
Metrics such as \textit{pass@$k$} and \textit{pass\textasciicircum{$k$}} extend this by considering success over multiple trials \cite{yao_-bench_2024}.

This crucial objective is applied across a wide range of LLM agent evaluation domains and benchmarks \cite{wang_survey_2024}. This includes tasks related to coding and software engineering, such as resolving GitHub issues (SWE-bench \cite{jimenez_swe-bench_2024}), scientific data analysis programming (ScienceAgentBench \cite{chen_scienceagentbench_2025}), reproducing research (CORE-Bench \cite{siegel_core-bench_2024}, PaperBench \cite{starace_paperbench_2025}), and interactive coding in apps (AppWorld \cite{trivedi_appworld_2024}). It is also extensively used for agents interacting with web environments, including general web navigation (BrowserGym \cite{chezelles_browsergym_2025}, WebArena \cite{webarena}, WebCanvas \cite{pan_webcanvas_2024}), multimodal web tasks (VisualWebArena \cite{koh_visualwebarena_2024}, MMInA \cite{zhang_mmina_2024}), and realistic time-consuming web tasks (ASSISTANTBENCH \cite{yoran_assistantbench_2024}).

\subsubsection{Output Quality:} Output quality refers to the characteristics of responses by an LLM agent. It is an umbrella term encompassing aspects such as accuracy, relevance, clarity, coherence, and adherence to agent specifications or task requirements \cite{reimann_predicting_2023}. An agent may complete a task yet still deliver a subpar user experience if the interaction lacks the qualities mentioned above. Output quality is particularly relevant in evaluating conversational agents, where user goals are often achieved over multiple turns. Many metrics in this category overlap with those used in large language model (LLM) evaluation. For example, the fluency metric is used to measure the degree to which the output of an LLM satisfies the conventions of natural language \cite{zhang2023llmevalpreliminarystudyevaluate}. The logical coherence metric focuses on the rigor in arguments \cite{zhang2023llmevalpreliminarystudyevaluate}. As LLM agents may utilize tools for retrieving grounding information and providing context-aware text answers, standard metrics used in retrieval-augmented generation (RAG) systems also apply. Such metrics include Response Relevance or Factual Correctness \cite{es_ragas_2025}.

\subsubsection{Latency \& Cost:} Latency is a critical aspect of agent behavior, especially in scenarios where users interact with agents synchronously. Long wait times can significantly degrade user experience and erode trust in the system. A commonly used metric in this context is \textit{Time To First Token} (TTFT), which measures the delay before a user sees the first token of an LLM's response in streaming mode. For use cases where the agent operates asynchronously, \textit{End-to-End Request Latency}—the time to receive the complete response—is often more relevant \cite{nvidia_metrics}.

While cost is not directly observable by end-users, it plays a crucial role in determining the practicality of deploying agents at scale. We include cost as a measure of an agent's monetary efficiency. It is typically estimated based on the number of input and output tokens, which directly correlate with usage-based pricing in most LLM deployments.

\subsection{Agent Capabilities}
Beyond external behavior, evaluations often target specific capabilities of LLM-based agents that enable their performance. Key aspects of this category include tool use, planning and reasoning, memory and context retention, and multi-agent collaboration. Evaluating these capabilities helps determine an agent’s strengths and weaknesses on a more granular level.

\subsubsection{Tool Use:} Tool use is a core capability for LLM-based agents, enabling them to retrieve grounding information, perform actions, and interact with external environments. In this survey, tool use involves invocation of a single tool and is interchangeable with function calling; more complex cases of determining tool sequences for complex tasks will be discussed in \ref{subsection:planning}. Recent advances have allowed  LLMs, such as ChatGPT-3.5 and beyond, to support function calling natively. These models can autonomously decide whether to invoke a function, select the appropriate one from a candidate set, and generate the required parameters. As a result, LLM agents can directly build on the functions of the underlying model, allowing many of the evaluation techniques originally developed for LLMs to use tools \cite{li_api-bank_2023}.

The evaluation of the tool's use involves answering several key questions. First, can the agent correctly determine whether tool invocation is necessary for a given task? If so, can it select the appropriate tool from a defined set of candidates? Once the tool is selected, the agent must be able to identify the correct parameters required by the tool and then generate appropriate values for each parameter to ensure proper execution. In cases where the candidate toolset is extensive, the agent may also need to retrieve the correct tool from a repository based on a natural language description of the task \cite{shen_taskbench_2024}.

Several metrics have been proposed to assess these abilities. \textit{Invocation Accuracy} \cite{liu_advances_2025} evaluates whether the agent makes the correct decision about whether to call a tool at all. \textit{Tool Selection Accuracy} measures whether the proper tool is chosen from a list of options. \textit{Retrieval Accuracy} focuses on whether the system can retrieve the correct tool from a larger toolset, often measured using rank accuracy $k$. For ranking-based evaluation, \textit{Mean Reciprocal Rank (MRR)} quantifies the position of the correct tool in the ranked list. In contrast, \textit{Normalized Discounted Cumulative Gain} (NDCG) reflects how well the system ranks all relevant tools\cite{liu_advances_2025}.

Parameter-related evaluation involves two aspects. The \textit{ parameter name F1 score} \cite{shen_taskbench_2024} measures the agent’s ability to identify the parameter names required for a given function correctly and then correctly assign values to them. While some evaluations rely on the correctness of abstract syntax trees (ASTs) to check if the tool call is syntactically valid, this approach may miss semantic errors, such as incorrect or hallucinated parameter values, especially for parameters constrained to enumerated types \cite{patil_gorilla_2023}. To address this limitation, recent work, such as the Gorilla paper, has proposed execution-based evaluation, in which the system runs the tool calls and assesses their outcomes, offering a more comprehensive and grounded assessment of tool use capability \cite{patil_gorilla_2023}.

\subsubsection{Planning and Reasoning:} \label{subsection:planning} Planning and reasoning are essential capabilities for LLM-based agents, especially in complex tasks that require multiple steps or making decisions under uncertainty. Planning involves selecting the correct set of tools in an appropriate order. At the same time, reasoning enables agents to make context-aware decisions, either ahead of time or dynamically during task execution \cite{huang_understanding_2024}. T-eval \cite{chen_t-eval_2024} formulated planning evaluation as comparing the set of predicted tools against a reference. Since tool order and dependency also matter, some benchmarks adopt graph-based representations and introduce metrics such as \textit{Node F1} for tool selection and \textit{Edge F1} or \textit{Normalized Edit Distance} for assessing tool invocation sequences and structural accuracy \cite{shen_taskbench_2024}.

In dynamic environments, agents often need to interleave planning and execution, adapting their actions in response to evolving context \cite{huang_understanding_2024}. This pattern is illustrated by the ReAct paradigm, where agents alternate between reasoning steps and tool usage \cite{yao_react_2023}. Evaluating such adaptive reasoning requires more than comparing static plans—it demands metrics that reflect decision-making in real time. The T-Eval framework \cite{chen_t-eval_2024} addresses this by introducing a reasoning metric that assesses how closely an agent’s predicted next tool call aligns with the expected one at each step. This captures the agent’s ability to make informed decisions when tool outputs are not known in advance. Similarly, AgentBoard \cite{ma_agentboard_2024} proposes the metric \textit{Progress Rate}, which compares the agent’s actual trajectory against the expected one, offering a fine-grained measure of how effectively the agent advances toward its goal.

When agents are instructed to plan in the form of generating complete multi-step programs, evaluation methods from code generation become relevant. Benchmarks like ScienceAgentBench compare the generated plans against annotated references using program similarity metrics \cite{chen_scienceagentbench_2025}. Additionally, the \textit{Step Success Rate} has been proposed to measure the percentage of steps in the generated plan that are successfully executed, providing a holistic view of planning quality during execution \cite{gioacchini_agentquest_2024}. 

\subsubsection{Memory and Context Retention:} A critical capability for long-running agents is the ability to retain information throughout many interactions and apply previous context to current requests. Guan et al. \cite{NEURIPS2024_guan_kong_zhong} categorize memory evaluation in multi-turn conversations by \textit{Memory Span} (how long information is stored) and \textit{Memory Forms} (how information is represented). For example, LongEval \cite{krishna2023longevalguidelineshumanevaluation} and SocialBench \cite{chen2024socialbenchsocialityevaluationroleplaying} are benchmarks that test an agent’s context retention in long dialogues (40+ turns). An agent might be given a conversation that spans dozens of exchanges and later asked questions that require recalling details from early in the conversation. Maharana et al. \cite{maharana2024evaluatinglongtermconversationalmemory} demonstrate evaluation with dialogues spanning hundreds of turns (600+ turns), and Li et al. \cite{optimus_1} introduce memory-enhanced evaluation techniques, tracking how well agents maintain consistency in long-horizon tasks. These evaluations often use synthetic or logged conversations as datasets, and metrics include \textit{Factual Recall Accuracy} or \textit{Consistency Score} (no contradictions between turns). Memory evaluation may also consider working memory for tool-using agents (i.e., whether the agent keeps track of intermediate results) and forgetting strategies (i.e., whether it appropriately forgets irrelevant details to avoid confusion). 

\subsubsection{Multi-Agent Collaboration:} Evaluating multi-agent collaboration in LLM-based systems requires different methodologies compared to traditional reinforcement learning-driven coordination \cite{chan2023chateval,li2024survey,talebirad2023multi}. Unlike conventional agents that rely on predefined reward structures, LLM agents coordinate through natural language, strategic reasoning, and decentralized problem-solving \cite{han2024llm, guo2024large}. These capabilities are crucial in real-world applications such as financial decision-making and structured data analysis, where autonomous agents must exchange information, negotiate, and synchronize decision-making processes efficiently \cite{liu2024autonomous, optimus_1}. Autonomous Agents for Collaborative Tasks \cite{liu2024autonomous} evaluates \textit{Collaborative Efficiency}, assessing how well multiple agents share responsibilities and distribute tasks dynamically.

\subsection{Reliability}
Reliability is a crucial objective, especially as LLM agents are considered for use in enterprise and safety-critical applications. It encompasses consistency, robustness to variations, and trustworthiness of the agents' outputs. Unlike task performance (which might measure best-case capabilities), reliability evaluation probes worst-case and average-case scenarios.

\subsubsection{Consistency:} Consistency refers to the stability of the output when the same task is repeated multiple times \cite{liu_advances_2025}. Since LLMs are inherently non-deterministic, LLM-based agents also exhibit variability in their behavior. For agents to be trusted in enterprise or other high-stakes applications, they must demonstrate consistent performance across repeated runs of the same task. A commonly used metric in this context is pass@k, which measures the probability that an agent succeeds at least once over k attempts. However, a stricter measure of consistency is whether the agent succeeds in all k attempts. This is formalized in the $\tau$-benchmark as the \textit{pass\textasciicircum{$k$}} metric, which better captures the consistency requirements of mission-critical deployments.

\subsubsection{Robustness:} Robustness refers to the stability of an agent’s output when faced with input variations or changes in the environment. To remain effective and trustworthy, LLM-based agents must consistently deliver high performance under a range of challenging conditions. Evaluating robustness often involves stress-testing the agent with perturbed inputs—such as paraphrased instructions, irrelevant or misleading context, or linguistic variations like typos and dialects—to assess whether it can still complete the task successfully. For example, robustness evaluations may involve applying systematic transformations to standard prompts and measuring the resulting drop in task success rate or output quality. The HELM benchmark \cite{liang_holistic_2023} explicitly incorporates such tests, tracking how model performance degrades under input variation.

Robustness also encompasses adaptive resilience—the agent’s ability to recover from dynamic changes in the environment. For instance, WebLinX \cite{Lu2024-gx} examines how agents behave when the structure of a web page changes during execution. In such settings, an effective agent must adjust its strategy rather than stall or fail.

In tool-using agents, robustness is further reflected in error-handling capabilities. As demonstrated in ToolEmu’s evaluation \cite{ToolEmu}, agents must be able to respond to tool failures or unexpected outputs gracefully. Robustness tests may include intentionally injecting failures—such as API errors or null responses—to observe whether the agent recovers (e.g., retries, switches tools, or explains the issue to the user) or breaks down. A key metric could be the proportion of induced failures that are handled appropriately, reflecting the agent's reliability in uncertain or imperfect conditions.

\subsection{Safety and Alignment} 
Safety covers an agent’s adherence to ethical guidelines, avoidance of harmful behavior, and compliance with legal or policy constraints. As LLM agents become more powerful and autonomous, the risk of unintended adverse outcomes (e.g., generating disinformation, hate speech, or unsafe instructions) grows, making safety evaluation indispensable. These evaluations are especially critical in fields such as financial services, cybersecurity, and autonomous decision-making, where agent vulnerabilities can lead to severe consequences \cite{li2024personal,fang2024llm,he2024emerged,gan2024navigating}. 

\subsubsection{Fairness:} The lack of fairness and transparency in AI agents can result in biased outcomes, decreased users' trust, and unintended societal consequences \cite{de2024can}. In financial applications, for example, biased decision-making in loan approvals or investment strategies can reinforce systemic inequalities (FinCon \cite{FinCon}, AutoGuide \cite{autoguide}). Ethical concerns also arise in multi-agent interactions, where decision-making frameworks must ensure compliance with standards and social norms \cite{mehrotra2024integrity}.

Explainability is crucial in enhancing user trust, especially in interactive systems where AI agents provide recommendations or automated assistance. Methods such as guideline-driven decision-making (AutoGuide \cite{autoguide}) and structured transparency mechanisms (MATSA \cite{matsa}, FinCon \cite{FinCon}) provide users with clear reasoning paths. Meanwhile, Rjudge \cite{Rjudge} analyzes how agents perceive risk when making autonomous decisions, emphasizing transparency and trustworthiness in AI interactions. Evaluating these dimensions ensures that AI agents align with ethical standards while maintaining fairness in their operational contexts.

\subsubsection{Harm, Toxicity, and Bias:} One aspect of safety is ensuring that an agent’s outputs do not contain harmful content such as hate speech, harassment, or extremely biased statements. Evaluation for toxicity often uses specialized test sets and metrics, such as the RealToxicityPrompts dataset \cite{gehman2020realtoxicitypromptsevaluatingneuraltoxic} – a collection of prompts likely to elicit toxic content—where its responses are checked with automated toxicity detectors or human raters. Metrics include the percentage of responses containing toxic language or the average toxicity score (as given by a classifier). HELM \cite{liang_holistic_2023} includes toxicity and bias metrics as part of holistic evaluation, indicating how frequently a model produces offensive content or exhibits undesired biases. For an interactive agent, one might evaluate it by giving provocative or ethically challenging inputs (red-teaming) and then measuring its failure rate (how often it responds in an unsafe manner). Safety-focused datasets, such as CoSafe, target exactly this: Yu et al. introduce CoSafe \cite{cosafe} to evaluate conversation agents on adversarial prompts designed to trick them into breaking safety rules (e.g., a user subtly asks for self-harm advice or illicit instructions). CoSafe revealed that even advanced agents had vulnerabilities, such as falling for coreference-based attacks (where a user refers to something ambiguously to bypass filters). The evaluation process involved monitoring the agent’s responses for policy violations when faced with these adversarial queries. Having a numeric score (like “agent produced a disallowed response in X\% of adversarial cases”) quantifies safety. 

\subsubsection{Compliance and Privacy:} Beyond avoiding overt toxicity, many deployments require agents to comply with specific regulatory or policy constraints \cite{zhang2023sa, brown2024enhancing}. For instance, a finance chatbot must not disclose confidential information or provide particular types of financial advice, and a medical assistant must not deviate from established medical guidelines. Evaluating compliance may be highly domain-specific, as it involves scenarios crafted to test whether the agent respects boundaries (e.g., a user asks the medical bot for a prescription-only drug recommendation—the correct, safe behavior is to refuse and advise consulting a doctor). 

In enterprise contexts, compliance evaluation may require proprietary test cases that reflect actual policies. One approach is to integrate those concerns into evaluation frameworks. For example, the HELM benchmark \cite{liang_holistic_2023} for enterprises was proposed to include domain-specific prompts and metrics (such as accuracy on financial jargon or compliance in responses) for fields like finance and law. The process involves gathering representative enterprise scenarios (which may contain confidential or custom data) and designing evaluation metrics that reflect real-world success criteria (e.g., Did the agent follow all legal disclaimer requirements in its response?). For example, TheAgentCompany \cite{theagentcompany} evaluates enterprise AI agents under structured correctness constraints, requiring them to follow predefined organizational policies when completing tasks. 



\begin{table*}[htb]
    \caption{Evaluation Objectives and Their Metrics.}
    \label{table:objectives_and_metrics}
    \renewcommand{\tabcolsep}{0.3pc} 
    \renewcommand{\arraystretch}{1.0} 
\begin{tabularx}{\linewidth}{|>{\hsize=.4\hsize}X|>{\hsize=0.5\hsize}X|>{\hsize=0.8\hsize}X|>{\hsize=1.3\hsize}X|}
            \hline
            \textbf{Objectives} & \textbf{Category} & \textbf{Metrics} & \textbf{Relevant Papers} \\ 
            \hline
            Agent Behavior & Task Completion & Success Rate (SR), F1-score, Pass@k, Progress Rate, Execution Accuracy, Transfer Learning Success, Zero-Shot Generalization Accuracy & AgentBoard \cite{agentboard}, WebShop \cite{webshop}, AgentBench \cite{agentbench}, Tool Use Evaluation \cite{tooluse}, InformativeBench, SQuAD \cite{squad}\cite{squad2} \cite{autonomousagents}, ResearchArena~\cite{ResearchArena}, AgentBoard~\cite{agentboard}, AppWorld \cite{trivedi_appworld_2024}, TheAgentCompany~\cite{theagentcompany}, MAGIC \cite{magic}, Mobile-Env \cite{mobileenv}, Re-ReST \cite{rerest}, XMC-AGENT \cite{xmcagent}, SWE-bench \cite{jimenez_swe-bench_2024}  \\ \cline{2-4}
                            & Output Quality & Coherence, User Satisfaction, Usability, Likability, Overall Quality & PredictingIQ \cite{reimann_predicting_2023}, EnDex \cite{endex}, PsychoGAT~\cite{Yang_Wang_Chen_2024} \\ \cline{2-4}
                            & Latency \& Cost & Latency, Token Usage, Cost & Cluster diagnosis~\cite{Shi2024}, MobileBench \cite{mobilebench}, MobileAgentBench \cite{mobileagentbench}, LangSuitE~\cite{langsuite}, WebArena~\cite{webarena}, Mobile-env \cite{zhang2023mobile}, GUI Agents \cite{wang2024gui}, GPTDroid \cite{liu2023chatting}, Spa-bench \cite{chen2024spa} \\ 
            \hline
            Agent Capability  & Tool Use & Task Completion Rate, Tool Selection Accuracy & ToolEmu \cite{Ruan2024}, MetaTool \cite{huang2024metatoolbenchmarklargelanguage}, AutoCodeRover \cite{autocoderover}  \\ \cline{2-4}
            & Planning \& Reasoning & Reasoning Quality, Accuracy, Fine-Grained Progress Rate, Self Consistency, Plan Quality & AgentBoard \cite{agentboard} Massive Multitask Language Understanding \cite{hendrycks2021measuringmassivemultitasklanguage}, LLM-Augmented Autonomous Agents \cite{liu2024autonomous}, Cluster diagnosis questions~\cite{Shi2024}, SimuCourt \cite{He_Cao_Wang_2024}, Magis \cite{tao2024magis} \\ \cline{2-4}
            & Memory \& Context Retention & Factual Accuracy Recall, Consistency Scores & LongEval \cite{krishna2023longevalguidelineshumanevaluation}, SocialBench \cite{chen2024socialbenchsocialityevaluationroleplaying}, LoCoMo \cite{maharana2024evaluatinglongtermconversationalmemory}, Optimus-1 \cite{optimus_1} \\ \cline{2-4}
            & Multi-Agent Collaboration & Information Sharing Effectiveness, Adaptive Role Switching, Reasoning Rating & AgentSims~\cite{lin2023agentsims}, WebArena \cite{webarena}, MATSA~\cite{matsa}, GAMEBENCH \cite{gamebench}, BALROG \cite{balrog}, TheAgentCompany \cite{theagentcompany}  \\ 
            \hline
            Reliability  & Consistency & pass\textasciicircum{k} & $\tau$-Bench \cite{yao_-bench_2024} \\ \cline{2-4}
            & Robustness & Accuracy, Task Success Rate Under Perturbation &  HELM \cite{liang_holistic_2023}, WebLinX \cite{Lu2024-gx} \\ \cline{2-4}
            \hline
            Safety  & Fairness & Awareness Coverage, Violation Rate, Transparency, Ethics, Morality & CASA~\cite{Qiu2025}, R-Judge~\cite{Rjudge}, SimuCourt~\cite{He_Cao_Wang_2024}, MATSA \cite{matsa}, FinCon \cite{FinCon}, AutoGuide \cite{autoguide} \\ \cline{2-4}
            
            & Harm & Adversarial Robustness, Prompt Injection Resistance, Harmfulness, Bias Detection & Agent Security Bench(ASB) \cite{agentsecbench}, AgentPoison \cite{agentpoison}, AgentDojo \cite{debenedetti_agentdojo_2024}, Backdoor Attacks \cite{backdoor}, SafeAgentBench \cite{safeagentbench}, Agent-Safety Bench \cite{agentsafetybench}, AgentHarm \cite{andriushchenko_agentharm_2025},  Adaptive Attacks \cite{zhan2025adaptiveattacksbreakdefenses}, RealToxicityPrompts \cite{gehman2020realtoxicitypromptsevaluatingneuraltoxic}\\ \cline{2-4}
            & Compliance \& Privacy & Risk Awareness, Task Completion Under Constraints & R-Judge \cite{Rjudge}, Cybench \cite{cybench}, TheAgentCompany \cite{theagentcompany} \\ 
            \hline
            
    \end{tabularx}\\[5pt]
\end{table*}

\section{Evaluation Process} 

\subsection{Interaction Mode}
Evaluating LLM agents can occur in various interaction modes and with different tooling. A fundamental distinction is between offline evaluation (using pre-generated, static datasets) and online assessment (involving reactive simulations, humans in the loop, or live system monitoring).  

\subsubsection{Static \& Offline Evaluation} Often performed as a baseline, offline evaluations typically rely on datasets and static test cases: collections of tasks, prompts, or conversations that represent challenges the agent might face. Simulated conversations may be used to help develop these data, but are ultimately inert between different runs. Though comparatively cheaper and simpler to run and maintain, offline evaluations typically lack the nuance to fully address the wide range of responses an LLM agent may be able or expected to provide. As such, they are more prone to error propagation and are generally less accurate representations of system performance.

\subsubsection{Dynamic \& Online Evaluation} As with other ML systems, online evaluation often occurs after an LLM agent has been deployed. Instead of relying on synthetic, historical, or manually crafted data, online evaluations leverage simulations or fundamental user interactions. This adaptive data is crucial for identifying pain points and issues that are not discovered during static testing and is often rich in domain context that is more difficult to capture with synthetic or generalized benchmarks. Dynamic evaluations may use proxies to \textbf{simulate users or environments} in reactive, real-time response to agent behavior. For example, in the assessment of web agents, researchers built web simulators (MiniWoB \cite{miniwob}, WebShop \cite{webshop}, WebArena \cite{webarena}, etc.) where the behavior of an agent (clicking links, filling forms) can be programmed to verify the correct sequence. 

The concept of \textbf{Evaluation-driven Development (EDD)} has also been proposed \cite{xia2024evaluation}, proposing making evaluation an integral part of the agent development cycle. It advocates for continuous evaluation of the agent, both offline (during development) and online (after deployment), to detect regressions and adapt to new use cases. They further outline a reference architecture in which an AgentOps component monitors agent performance in production and provides insights back to developers. While still an emerging idea, it underscores that evaluation is not a one-time task but an ongoing process, especially for agents that learn or evolve.

\subsection{Evaluation Data} The growing interest in evaluating LLM-based agents has led to the development of diverse datasets, benchmarks, and leaderboards specifically targeting the agent capabilities discussed in Section \ref{sec:eval_objectives}. Many of these resources are designed to reflect real-world complexity and are built using a mix of human-annotated, synthetic, and interaction-generated data. For instance, datasets such as AAAR-1.0 \cite{lou_aaar-10_2025}, ScienceAgentBench \cite{chen_scienceagentbench_2025}, and TaskBench \cite{shen_taskbench_2024} provide structured, expert-labeled benchmarks for assessing research reasoning, scientific workflows, and multi-tool planning. Others, such as FlowBench \cite{xiao_flowbench_2024}, ToolBench \cite{huang2024metatoolbenchmarklargelanguage}, and API-Bank \cite{li_api-bank_2023}, focus on tool use and function-calling across large API repositories. These benchmarks typically include not only the gold tool sequences but also expected parameter structures, enabling fine-grained evaluation.

In parallel, datasets like AssistantBench \cite{yoran_assistantbench_2024}, AppWorld \cite{trivedi_appworld_2024}, and WebArena \cite{webarena} simulate more open-ended and interactive agent behaviors in web and application environments. They emphasize dynamic decision-making, long-horizon planning, and user-agent interactions. Several benchmarks also support safety and robustness testing—for example, AgentHarm \cite{andriushchenko_agentharm_2025} assesses potentially harmful behaviors, while AgentDojo \cite{debenedetti_agentdojo_2024} evaluates resilience against prompt injection attacks. Leaderboards such as the Berkeley Function-Calling Leaderboard (BFCL) \cite{bfcl} and Holistic Agent Leaderboard \cite{stroebl_hal_2025} consolidate these evaluations by providing standardized test cases, automated metrics (e.g., AST correctness, Win Rate), and ranking mechanisms to compare systems. 

\subsection{Metrics Computation Methods}
The code-based method is the most deterministic and objective approach \cite{agentboard, liu_agentbench_2023, tooluse}. \cite{agentboard} It relies on explicit rules, test cases, or assertions to verify whether an agent’s response meets predefined criteria. This method is particularly effective for tasks with well-defined outputs, such as numerical calculations, structured query generation, or syntactic correctness in programming tasks. Its primary advantage is its consistency and reproducibility, making it highly reliable for benchmarking. However, code-based methods are typically inflexible. They struggle with evaluating open-ended or qualitative responses, such as natural language generation or creative problem-solving, where correctness is subjective. Despite this, it remains a fundamental technique for evaluating structured tasks where correctness is well-defined.

The LLM-as-a-Judge approach \cite{zheng2023judging} leverages the reasoning capabilities of LLMs to evaluate agent responses based on qualitative criteria, assessing responses according to the criteria provided through instructions. This method has gained traction due to its ability to handle tasks that are subjective and nuanced, such as summarization, reasoning, and conversational interactions. One recent extension of this method is Agent-as-a-Judge \cite{zhuge2024agent}, where the evaluation process involves multiple AI agents interacting to refine the assessment, potentially improving evaluation reliability. This method is highly scalable and can adapt to complex tasks. As a result, it has received increasing attention \cite{li2024llms, gu2024survey}. 

Human-in-the-loop evaluation remains the gold standard for subjective aspects (like naturalness and user satisfaction) and safety-critical judgment calls. Human evaluations can take the form of user studies, expert audits (where domain experts review agent outputs), or Crowdworker annotations (where outputs are rated along dimensions such as relevance, correctness, and tone). This method offers the highest reliability in open-ended tasks, such as content generation, strategic decision-making, or dialogue coherence. However, it is expensive, time-consuming, and challenging to scale, making it impractical for large-scale automated systems that require frequent evaluations. 

\subsection{Evaluation Tooling}
A notable aspect in the process dimension is the emergence of software frameworks and platforms that support automated, scalable, and continuous agent evaluation workflows. These tools enable integration of evaluation directly into the development lifecycle, reflecting a growing movement toward Evaluation-driven Development (EDD) \cite{xia2024evaluation} in agent building. OpenAI Evals is an open-source framework that allows developers to specify evaluation tasks and metrics for models, automating the execution and reporting of results (though not formally described in academic literature, it reflects practical needs) \cite{openai_eval}. Other open-source or commercial tools such as DeepEval \cite{deepeval}, InspectAI \cite{inspectai}, Phoenix \cite{phoenix_eval}, and GALILEO \cite{galileo2025agentic} provide rich analytics, evaluation orchestration, and debugging capabilities. Moreover, agent development platforms like Azure AI Foundry \cite{azure_foundry}, Google Vortex AI \cite{google_vortex_ai}, LangGraph \cite{langgraph}, and Amazon Bedrock \cite{amazon_bedrock} increasingly incorporate evaluation features, helping developers monitor performance, detect regressions, and adapt agents to evolving user needs. Xia et al. \cite{xia2024evaluation} further propose an AgentOps architecture to continuously monitor deployed agents, closing the loop between development and deployment through real-time feedback and quality control.

\subsection{Evaluation Contexts}
The evaluation context pertains to the environment in which an evaluation is performed. Similar to software engineering, a tradeoff exists between more realistic (but often more costly and potentially less secure) and simpler, more controlled (but usually less representative of final performance) environments. The context in which a system is assessed is typically guided by the system's intended use; a simpler LLM agent without edit access might be tested in its working environment directly, whereas an LLM agent designed to work with and make changes to many intertwined systems will likely be evaluated in a mocked API or sandbox environment. For less contained systems, the evaluation context may take the form of a web simulator, such as MiniWoB \cite{miniwob}, WebShop \cite{webshop}, or WebArena \cite{webarena}. As the development of an agent continues, the evaluation context often evolves with it, from smaller, mocked API environments to live deployment as agent performance and trustworthiness are determined.


\section{Enterprise-Specific Challenges}
As LLM-based agents transition from research demos to deployment in enterprise settings, new challenges are emerging. Enterprises often demand high performance in conjunction with predictable reliability, compliance with regulations, data security, and maintainability, which are usually overlooked during evaluation. To address these gaps, we discuss the concerns outlined in the following sections and outline future directions. 

\subsection{Complexity from Role-based Access}
A key challenge in evaluating LLM-based agents in enterprise settings is the need to account for Role-Based Access Control (RBAC), which governs users' permissions to access data and services. In these environments, users operate with varying levels of access depending on their roles, and agents acting on their behalf must adhere to the same constraints. This introduces complexity into agent evaluation, as an agent’s ability to retrieve or act on information is not uniform but contextually bound to the user’s permissions. 

To address this, some evaluation frameworks have begun incorporating access control constraints into their design. For example, IntellAgent \cite{levi_intellagent_2025} includes evaluation tasks that require authentication of user identity and enforce policies that deny access to other users' information. By embedding role-specific restrictions into task generation, these approaches more accurately model how agents behave in permission-sensitive enterprise contexts.

\subsection{Reliability Guarantees} 
Reliability guarantees are especially important in enterprise settings, where agents are expected to operate within compliance and auditing frameworks that require deterministic or repeatable behavior that is explainable. In such contexts, occasional success is insufficient; agents must perform reliably across time and usage scenarios to be considered production-ready.

Evaluating reliability is nontrivial. Because LLM-based agents are inherently stochastic, measuring consistency requires executing the same task multiple times and observing the variation in outcomes. This introduces significant evaluation overhead: running multiple trials per input can be computationally expensive, especially when testing complex tasks involving tools, memory, or multi-agent coordination. Moreover, to draw meaningful conclusions, benchmarks must include a representative dataset that reflects the types of tasks and conditions the agent may encounter.

Some efforts have begun to tackle this challenge. For example, the $\tau$-benchmark \cite{yao_-bench_2024} explicitly incorporates the pass\textasciicircum{$k$} metric to evaluate the consistency of an agent. By applying this to domains such as retail and airline booking, the authors show that current agents struggle with consistency.

\subsection{Dynamic and Long-Horizon Interactions} 
One major challenge in evaluating LLM-based agents is assessing their performance on long-horizon tasks in dynamic, evolving environments. Unlike most current benchmarks that focus on short episodes or single interactions, real-world enterprise agents often operate continuously over extended periods while interacting with users, systems, and data.

Addressing this challenge is essential for understanding how agents behave over time, particularly in enterprise settings where reliability, adaptability, and goal alignment are crucial throughout the agent’s lifecycle. Standard, short-term evaluations cannot capture phenomena such as performance drift, context retention, or the cumulative effect of decisions on business outcomes.

To begin tackling this issue, some research efforts have introduced long-running simulations and extended dialogues as evaluation tools. For instance, Park et al. \cite{park2023generative} observed generative agents in a continuously running simulated town environment to study emergent behaviors across multi-day interactions. Similarly, Maharana et al. \cite{maharana2024evaluatinglongtermconversationalmemory} evaluated long-term conversational memory through 600-turn dialogues, focusing on how well agents maintain coherence and context over extended conversations. 

\subsection{Adherence to Domain-Specific Policies and Compliance Requirements} 
Another significant challenge in evaluating enterprise agents is ensuring that they can operate within domain-specific policies and compliance constraints. Enterprises often enforce strict operational rules—such as approval workflows, data retention policies, usage quotas, and legal regulations like GDPR or HIPAA—that must be respected by agents throughout task execution. Evaluating agents in such contexts requires more than measuring task success; it demands verification that agent behaviors align with formal policy constraints and legal compliance standards. For instance, an agent generating financial reports must avoid unauthorized access to confidential forecasts and ensure that generated content adheres to regulatory reporting standards. Without explicit modeling of these constraints during evaluation, agents deemed "correct" in traditional benchmarks may still fail in production due to policy violations or compliance risks.

\section{Future Research Directions}
As LLM-based agents continue to grow in complexity and application scope, future research must push toward more robust, practical, and scalable evaluation methodologies. We highlight four key directions that can significantly advance the field:

\textbf{Holistic Evaluation Frameworks}: Current evaluation efforts often focus on isolated dimensions such as task success, planning quality, or tool use. However, agents in real-world applications must simultaneously balance multiple competencies. Future work should develop holistic evaluation frameworks that assess agent performance across multiple, interdependent dimensions.

\textbf{More Realistic Evaluation Settings:} To bridge the gap between lab settings and production environments, agent evaluation must move toward more realistic conditions. This includes creating evaluation environments that incorporate enterprise-specific elements such as dynamic multi-user interactions, role-based access controls, and domain-specific knowledge. These settings can be achieved through real-world deployment trials or through simulated environments that mimic enterprise workflows. 

\textbf{Automated and Scalable Evaluation Techniques}: Manual evaluation of agent behavior, especially in multi-turn or long-horizon scenarios, is costly and complicated to scale. Future research should explore automated evaluation approaches to reduce human effort and improve reproducibility. This includes using synthetic data generation for controllable test cases, leveraging simulated environments to emulate task contexts, and advancing LLM-based evaluation techniques such as LLM-as-a-judge or Agent-as-a-judge.

\textbf{Time- and Cost-Bounded Evaluation Protocols}: Evaluation must be efficient and able to support iterative agent development. Today’s methods—especially those requiring repeated trials or human-in-the-loop assessments—can be both time- and resource-intensive. Future research should aim to develop time- and cost-bounded evaluation protocols that strike a balance between depth and efficiency. 

In summary, future research should focus on developing evaluation methods that are holistic, realistic, scalable, and efficient. These directions are essential for building reliable and trustworthy LLM-based agents at scale.

\bibliographystyle{ACM-Reference-Format}
\balance
\bibliography{references}

\end{document}